\pdfoutput=1

\documentclass[11pt]{article}

\usepackage{ACL2023}

\usepackage{times}
\usepackage{latexsym}

\usepackage[T1]{fontenc}

\usepackage[utf8]{inputenc}

\usepackage{microtype}

\usepackage{inconsolata}

\usepackage{amsmath}
\usepackage{float}
\usepackage{mdframed}
\usepackage{listings}
\usepackage{graphicx}
\usepackage{multirow}
\usepackage{booktabs}
\usepackage{url}

\title{Language Models can perform Single-Utterance Self-Correction of Perturbed Reasoning}

\author{Sam Silver \\
  \texttt{UC Santa Cruz} \\\And
  Jimin Sun \\
  \texttt{Cohere, CMU} \\\And
  Ivan Zhang \\
  \texttt{Cohere} \\\And
  Sara Hooker \\
  \texttt{Cohere} \\\And
  Eddie Kim \\
  \texttt{Cohere}
  }

\begin{document}
\maketitle
\begin{abstract}
Large Language Models (LLMs) have demonstrated impressive mathematical reasoning capabilities, yet their performance remains brittle to minor variations in problem description and prompting strategy. Furthermore, reasoning is vulnerable to sampling-induced errors which autoregressive models must primarily address using self-correction via additionally-generated tokens. 
To better understand self-correction capabilities of recent models, we conduct experiments measuring models' ability to self-correct synthetic perturbations introduced into their Chain of Thought (CoT) reasoning. 
We observe robust \textit{single-utterance} intrinsic self-correction behavior across a range of open-weight models and datasets, ranging from subtle, implicit corrections to explicit acknowledgments and corrections of errors. 
Our findings suggest that LLMs, including those not finetuned for long CoT,  may possess stronger intrinsic self-correction capabilities than commonly shown in the literature. The presence of this ability suggests that recent "reasoning" model work involves amplification of traits already meaningfully present in models.
\end{abstract}

\section{Introduction}
Large Language Models (LLMs) have shown progressively impressive performance in mathematical domains (\citealp{cobbe2021trainingverifierssolvemath}; \citealp{hendrycks2021measuring}; \citealp{lewkowycz2022solving}; \citealp{yang2024qwen25mathtechnicalreportmathematical}), owing largely to improvements in data curation and post-training techniques.

At inference time, researchers have found  that performance can be substantially improved by encouraging models to generate natural language rationales that allow for an adaptive amount of computation for each subproblem (\citealp{nye2022show};\citealp{NEURIPS2022_9d560961}; \citealp{zhou2023least}; \citealp{zheng2024step}, \textit{inter alia}). 

However, despite the apparent sophistication of LLM reasoning capabilities, recent work has documented a variety of reasoning failure modes.
For example, models have a tendency to fall into poorly performing reasoning patterns when presented with familiar but subtly modified problems \citep{mirzadeh2024gsmsymbolicunderstandinglimitationsmathematical}, can be easily distracted with irrelevant context \citep{shi2023large}, and are brittle to changes in premise ordering \citep{chen2024premise}.
Critically, LLMs struggle to identify their own errors and contradictions, making it difficult to trust outputs without external verification.

While a three-turn generate-critique-correct process with optimized prompting is popular in self-correction literature \cite{NEURIPS2023_91edff07}, recent trends in frontier language model releases (\citealp{openai2024o1}; \citealp{qwq2024reflect}; \citealp{pichai2024gemini}; \citealp{deepseek2025incentivizing}) point to a growing interest in models' ability to perform self-evaluation \textit{intrinsically} at test-time in a \textbf{single-utterance}, without aid from external verifiers. A critical component of this behavior is intrinsic self-correction, when models recognize an error in their reasoning, acknowledge the mistake, and output a corrected generation.

To better understand current capabilities around single-utterance intrinsic self-correction, we introduce a novel experimental framework focused on evaluating how LLMs recover from perturbations in their reasoning chains. Our results
reveal that language models, even those not trained as "reasoning" models, can successfully recover from introduced reasoning perturbations, exhibiting both implicit and explicit self-correction behavior.

\begin{figure*}[t]
    \centering
    \includegraphics[width=1\linewidth]{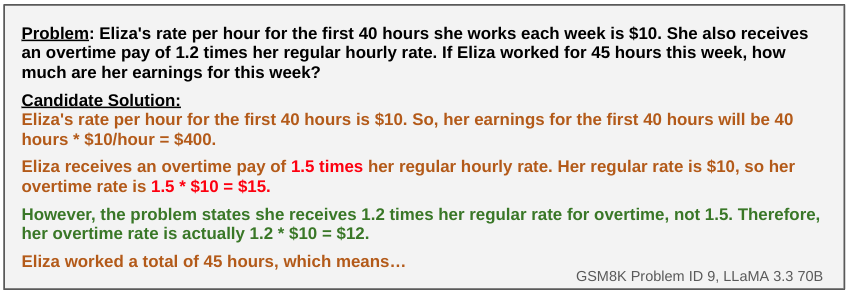}
    \caption{Truncated excerpt of a candidate solution (brown) showing LLaMA 3.3 70B explicitly self-correcting (green) mid-generation during single-utterance completion of a perturbed (red) on-policy reasoning stub.}
    \label{fig:llama-short-recovery}
\end{figure*}

\section{Related Work}
In contrast to approaches that rely on external feedback (See Appendix~\ref{subsec:extrinsic-feedback}), recent work has explored methods to enable LLM self-correction using only their own parametric knowledge. 

Prompt-based self-correction techniques involve models reviewing and revising their own outputs, checking for potential errors, inconsistencies, or misalignment (\citealp{bai2022constitutionalaiharmlessnessai}; \citealp{saunders2022selfcritiquingmodelsassistinghuman}).
These self-refinement processes can be iterated, allowing for rounds of reflection and refactoring to improve responses (\citealp{NEURIPS2023_91edff07}; \citealp{ye2023selfee}). \citet{yuan2024self} applies a similar iterative strategy where self-critique takes the form of the generator itself acting as a judge of its own responses, using a rubric and its own judgment to assign a scalar reward to generations.

Other approaches aim to develop models that robustly recognize and correct their own errors at a level beyond that offered by simple prompting by incorporating self-correction training into model training \citep{kumar2024traininglanguagemodelsselfcorrect}.

The most recent and emerging advances in intrinsic self-evaluation focus on single-utterance techniques in which models continuously monitor, assess, and refine their generation trajectories (\citealp{openai2024o1}; \citealp{qwq2024reflect}). 
\citet{lambert2024tulu3pushingfrontiers}, \citet{deepseek2025incentivizing}, and \citet{kimi2025scaling} have offered concrete insights into how simple reinforcement learning (RL) against verifiable outcomes effectively elicits improved reasoning performance and qualitatively similar generation styles. In particular, DeepSeek's R1-Zero highlights that self-evaluating behavior can be elicited directly from high-quality \textit{base} models, and that this behavior can be distilled into models as small as 1.5B parameters \citep{deepseek2025incentivizing}.

Still, there exists criticism of self-correction experiments as commonly-performed in literature: \citet{huang2024large} present perhaps the most direct challenge to the optimism surrounding self-correction capabilities, finding that language models, in a three-turn generate-critique-correct pipeline, not only struggle to reliably correct their own reasoning, but often perform \textit{worse} after attempting intrinsic self-correction in a setting in which helpful information and criteria are not imparted into the critique prompt.

In contrast to either the extrinsic feedback approaches or multi-turn prompt-based intrinsic correction approaches, we examine models' ability to perform single-utterance intrinsic self-correction of introduced perturbations. 

\begin{figure*}[t]
    \centering
    \includegraphics[width=1\linewidth]{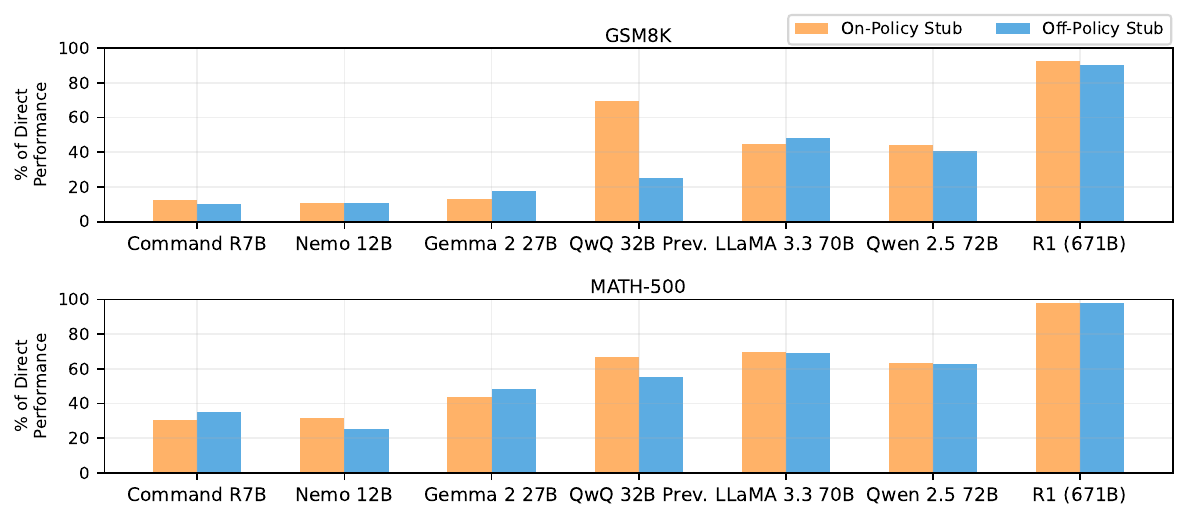}
    \caption{Perturbation recovery success rate in on-policy (orange) and off-policy (blue) reasoning stub scenarios shown for GSM8K and MATH-500 as a percentage of the success rate of the direct, unperturbed scenario. A 100\% represents equivalent performance to the unperturbed scenario.}
    \label{fig:relative-recovery-rates}
\end{figure*}

\section{Experiments}
\label{sec:experiments}

To better understand self-correction capabilities in language models, we designed an experimental framework to measure the elicitation of intrinsic self-correction under synthetically perturbed reasoning trajectories. We evaluate a variety of models' recovery performance in the context of popular math reasoning datasets (See Appendix~\ref{subsec:dataset-description}).

For each model, our approach involves four phases detailed in Figure~\ref{fig:experiment-diagram} and Appendix~\ref{sec:prompts}: 
First, each evaluated model is prompted with a reasoning problem and generates a 100-token solution "stub." These stubs empirically contain enough progress to enable effective perturbation, but not so much as to leave no headroom for recovery.

Next, a held-out model (LLaMA 3.1 405B) applies a reasoning perturbation to the solution stub. Perturbations include changing decimal places (e.g. from "1.5" to "15"), switching operators (e.g. from $\times$ to $\div$), altering a key phrase (e.g. from "60\% of \$5" to "60\% more than \$5"), or a number of other perturbations (shown in Figure~\ref{fig:perturbation-prompt-1}) similar to those used in \citet{sun2024toolsfaildetectingsilent}.

Then, the model under evaluation completes the generation stemming from the perturbed reasoning stub to finish the candidate solution. This stub generation and completion are seen from the model's perspective a single, uninterrupted utterance. 

Finally, a grader model (LLaMA 3.1 405B) with access to the ground-truth solution determines candidate solution correctness.

\subsection{Models and Datasets}

We evaluate seven modern language models of various size and origin, aiming to cover a range of sizes across diverse model families. Models include Command R7B \cite{cohere2024command}, Nemo 12B \citep{mistral2024nemo}, Gemma 2 27B \citep{gemmateam2024gemma2improvingopen}, QwQ 32B Preview \citep{qwq2024reflect}, LLaMA 3.3 70B \citep{grattafiori2024llama3herdmodels}, Qwen 2.5 72B \citep{qwen2024qwen25technicalreport}, and R1 \citep{deepseek2025incentivizing}. QwQ and R1 are advertised as "reasoning" models.
For perturbation generation and solution verification, we employ LLaMA 3.1 405B. See  Table~\ref{tab:providers} for more information on  inference providers and model precision.

These models are evaluated on three popular math reasoning datasets: GSM8K \citep{cobbe2021trainingverifierssolvemath}, GSM-Symbolic \cite{mirzadeh2024gsmsymbolicunderstandinglimitationsmathematical}, and MATH-500, a subset of the popular competition math dataset \cite{hendrycks2021measuring} as employed in \citet{lightman2023letsverifystepstep}. We additionally re-evaluated on the subset of the GSM8K dataset from which our GSM-Symbolic template-swapped sample was derived, which we refer to as "GSM8K Matched" (See Appendix~\ref{subsec:dataset-description}). 

\begin{table*}[!t]
\centering
\small
\setlength{\tabcolsep}{4pt}  
\begin{tabular}{lccc|ccc|ccc|ccc}
\toprule
\multirow{2}{*}{Model} & \multicolumn{3}{c|}{GSM8K} & \multicolumn{3}{c|}{GSM-Symbolic} & \multicolumn{3}{c|}{GSM8K Matched} & \multicolumn{3}{c}{MATH-500} \\
& Direct & Off & On & Direct & Off & On & Direct & Off & On & Direct & Off & On \\
\midrule
Command R7B & 88.0 & 8.8 & 10.8 & 85.0 & 8.0 & 12.0 & 94.0 & 7.0 & 9.0 & 59.0 & 20.6 & 18.0 \\
Nemo 12B & 87.7 & 9.2 & 9.3 & 87.0 & 10.0 & 6.0 & 89.0 & 8.0 & 7.0 & 45.4 & 11.4 & 14.4 \\
Gemma 2 27B & 90.8 & 16.3 & 12.1 & 94.0 & 22.0 & 11.0 & 90.0 & 14.0 & 11.0 & 57.8 & 28.0 & 25.4 \\
QwQ 32B Preview& 95.2 & 24.0 & 66.4 & 95.0 & 30.0 & 70.0 & 94.0 & 24.0 & 66.0 & 85.5 & 50.8 & 61.6 \\
LLaMA 3.3 70B & \textbf{96.4} & 46.4 & 43.1 & 94.0 & 54.0 & 41.0 & \textbf{96.0} & 48.0 & 40.0 & 75.0 & 51.8 & 52.2 \\
Qwen 2.5 72B & 95.1 & 39.0 & 41.8 & 93.0 & 45.0 & 46.0 & 95.0 & 43.0 & 39.0 & 85.1 & 53.8 & 54.0 \\
R1 (671B) & \textbf{96.4} & \textbf{87.0} & \textbf{89.3} & \textbf{98.4} & \textbf{76.8} & \textbf{80.0} & 95.0 & \textbf{88.0} & \textbf{90.0} & \textbf{91.9} & \textbf{90.0} & \textbf{90.0} \\
\bottomrule
\end{tabular}
\caption{Completion recovery success rate shown across datasets. \textit{Direct}, \textit{On}, and \textit{Off} refer to our Direction Solution, perturbed on-policy reasoning stub, and perturbed off-policy reasoning stub scenarios, respectively. Models range in size from 7B to 671B and are ordered by parameter count, ascending.}
\label{tab:results}
\end{table*}

\subsection{Evaluation Scenarios}
We evaluated models across three scenarios:

\textbf{Direct Solutions}: 
We evaluate each model's natural, unperturbed "pass@1" rate, giving a single opportunity to correctly solve each problem as determined by a grader language model with access to the ground-truth solution.

\textbf{Perturbed On-Policy Reasoning Stub}: 
We perform the four-phase workflow described in Section~\ref{sec:experiments} using our evaluated model to generate a reasoning stub which is then perturbed and completed.

\textbf{Perturbed Off-Policy Reasoning Stub}: We perform a similar set of experiments using a held-out language model to produce common reasoning stubs in the same four-phase workflow described in Section~\ref{sec:experiments}. This controlled set of perturbed stubs are individually completed by all evaluated models for an apples-to-apples comparison.

Model correction performance is evaluated using a simple accuracy metric, with the success rate $S$ computed as $S_a = \frac{c}{N}$, where $c$ is the number of correct solutions as determined by our grader language model and $N$ is the number of problems in the dataset. Models are accessed via OpenRouter or Cohere APIs in their original precision. Solution generation uses Top-P sampling \cite{holtzman2020curious} with P=0.8 and T=0.2, while perturbation and verification use greedy decoding.

\subsection{Results}
Our results demonstrate that self-correction capabilities are found across all evaluated models in the context of synthetic reasoning perturbations, even those not explicitly advertised as having been trained in single-utterance self-correction. Table~\ref{tab:results} shows accuracy performance across all datasets, scenarios, and models. Several key findings emerged from our experiment:

First, all models other than R1 experience meaningful performance degradations when errors are synthetically introduced into their reasoning process. We observe an average absolute drop in success rates (excluding R1) of 61.6\% in the on-policy scenario, with the smallest models suffering the largest relative performance drops.

We observe that smaller models (< 30B) experience a larger average drop of 78.1\% in absolute success rates, while larger models (>30B) experience a more modest drop of 41.7\% (ignoring R1). 
Qwen 2.5 72B and LLaMA 3.3 70B show surprisingly robust self-correction capabilities relative to QwQ 32B Preview and R1's impressive "reasoning" model performances, approximately matching the recovery performance of QwQ on MATH-500.

Recovery performance on GSM-Symbolic and MATH-500 is consistent with GSM8K results, suggesting that observed self-correction capabilities are not dataset-specific but rather indicative of general model ability. Interestingly, Figure~\ref{fig:relative-recovery-rates} shows that higher relative recovery rates were observed in the more difficult MATH dataset than in GSM8K.

QwQ exhibited degraded performance in the scenario involving completion of a perturbed off-policy reasoning stub. Examination of these completions indicates that QwQ's ability to initiate its characteristic self-evaluating style of generation is contingent on the style of the off-policy stub that it continues generation from.
This drop in off-policy performance suggests that the reasoning capability induced by QwQ's reinforcement learning finetuning may couple style with capability, with performance degrading when generating outside a familiar format distribution. In contrast, we observe R1 to be much more capable of re-initiating effective reasoning regardless of the reasoning stub's origin.

Finally, we observe a diversity of styles of self-correction behaviors on display from non-"reasoning" models, ranging from implicit correction behavior to explicit, well-aligned corrections as seen in Appendix~\ref{sec:examplerecoveries}. Explicit self-correction examples from non-"reasoning" models include "Wait a minute, let me double-check that because I think I might have made a mistake" and "However, the problem states that the discount is 30\%, not 50\%. Let's correct this and recalculate." We observe common use of critical "pivot tokens" (e.g. "Wait," "However," "Hold on") during generation in a manner reminiscent of the meta-cognitive "aha moment" highlighted in \citeposs{deepseek2025incentivizing} R1 technical report. 

The meaningful presence of these behaviors in our limited experiment suggests that strong models inherently possess latent self-correction capabilities, helping to explain why recent RL techniques have been particularly effective in promoting and amplifying these patterns.

\subsection{Conclusion}
Our work reveals that current language models may exhibit intrinsic self-correction capabilities more frequently than commonly believed, demonstrating that models can, in a single utterance, detect and recover from errors in their own reasoning chains without explicit prompting or external verification.

However, important limitations remain, as models commonly fail to detect simple introduced errors. Looking ahead, we believe that a better understanding of self-correction capabilities, more investigation into the coupling of style and reasoning in recent models, and improved methods for eliciting such behavior are crucial to developing reliable and trustworthy systems.

\newpage

\section{Limitations}
There are a variety of limitations of our current analysis that could be explored in future work.

\textbf{Off-Policy Perturbations}: We use a language model to apply perturbations to the reasoning of a model under evaluation. These errors that are introduced are likely to be significantly off-policy with respect to the models under evaluation, potentially making the recovery task artificially simple. Our reliance on API-based inference results in an inability to observe token-level probabilities at critical decision points in the reasoning process. If models were self-hosted, we could generate more realistic perturbations by selecting high-probability but incorrect continuations, creating a more natural experiment. 

\textbf{Dataset Coverage}: While we evaluated model recovery on multiple math datasets of varying difficulty, our analysis could benefit from the inclusion of even more challenging math and reasoning benchmarks. An earlier incarnation of our experiment tested recovery on the NuminaMath-CoT \citep{numina_math_datasets} and ZebraLogic \citep{lin2025zebralogicscalinglimitsllms} datasets through a different method of error introduction, but the results were inconclusive. Our current experimental setup could similarly be used to evaluate the robustness of model alignment by introducing misaligned perturbations to assistant responses or to evaluate model instruction-following abilities by violating stated constraints.

\textbf{Perturbation Abstraction}: Our perturbation methodology introduces errors that may be relatively easy for models to detect and correct. Rather than applying low-level perturbations like the corruption of arithmetic operations, future work could consider higher-level perturbations that significantly effect the problem-solving \textit{trajectory} of the model under evaluation. Such perturbations would be useful in evaluating a model's ability to perform reasoning backtracking.

\textbf{Assistant Prefill}: While we used a battery of heuristic prefill-completion tests to select OpenRouter model/provider combinations that seem to support the assistant prefill feature, OpenRouter and downstream inference provider documentation and support for this uncommonly-used feature is lacking, and we cannot guarantee with certainty that the assistant-prefill feature functions as advertised for each model/provider combination. 

\textbf{Model Availability}: Several promising open-weight models including DeepSeek 2.5 and the recently-released Deepseek V3 could not be evaluated due to the lack of inference provider support for the assistant prefill feature required for models to complete assistant turns prefixed by perturbed reasoning stubs. Similarly, many frontier closed-source models do not expose this feature.

\textbf{Provider Reliability}: We encountered reliability issues with certain model-provider combinations, particularly with QwQ 32B Preview and R1, leading to a small number of absent responses for cases in which 20 retries failed to yield 2XX responses. Although these data collection gaps are relatively small and do not meaningfully effect the results, they highlight the challenges of conducting large-scale evaluations using third-party inference providers.

\textbf{Scale Effects}: Our study does not systematically explore how self-correction capability varies with model scale within the same model family. While we observe positive correlation between model size and self-correction performance, a more controlled study across model scales within model families would be needed to draw stronger conclusions about whether explicit self-correction is emergent with model scale.

\textbf{Taxonomies}: We have not developed a comprehensive taxonomy of either perturbation types or observed correction strategies. Analysis powered by a more detailed categorization of both the kinds of errors introduced and the methods models use to recover could provide insights to improve model robustness.

\section{Ethics Statements}
Our research on self-correction capabilities in language models touches on several important ethical considerations.

\textbf{Reliability and Trust}: Understanding how language models detect and correct their own errors is crucial for developing more reliable AI systems, and especially critical in high-stakes applications where unchecked errors in reasoning could have serious social consequences.

\textbf{Dual Use Considerations}: Our work aims to improve our understanding of model self-correction capabilities, but enhancements in self-correction capabilities could improve LM-powered systems designed with nefarious intentions. We acknowledge the dual-use nature of AI systems and their potential for misuse.

\bibliography{cohere}
\bibliographystyle{acl_natbib}

\appendix

\onecolumn

\section*{\Large Appendices}

\section{Tables and Figures}

\renewcommand{\thefigure}{A\arabic{figure}}
\renewcommand{\thetable}{A\arabic{table}}
\setcounter{figure}{0}  
\setcounter{table}{0}   

\begin{figure}[H]
    \centering
    \includegraphics[width=1\linewidth]{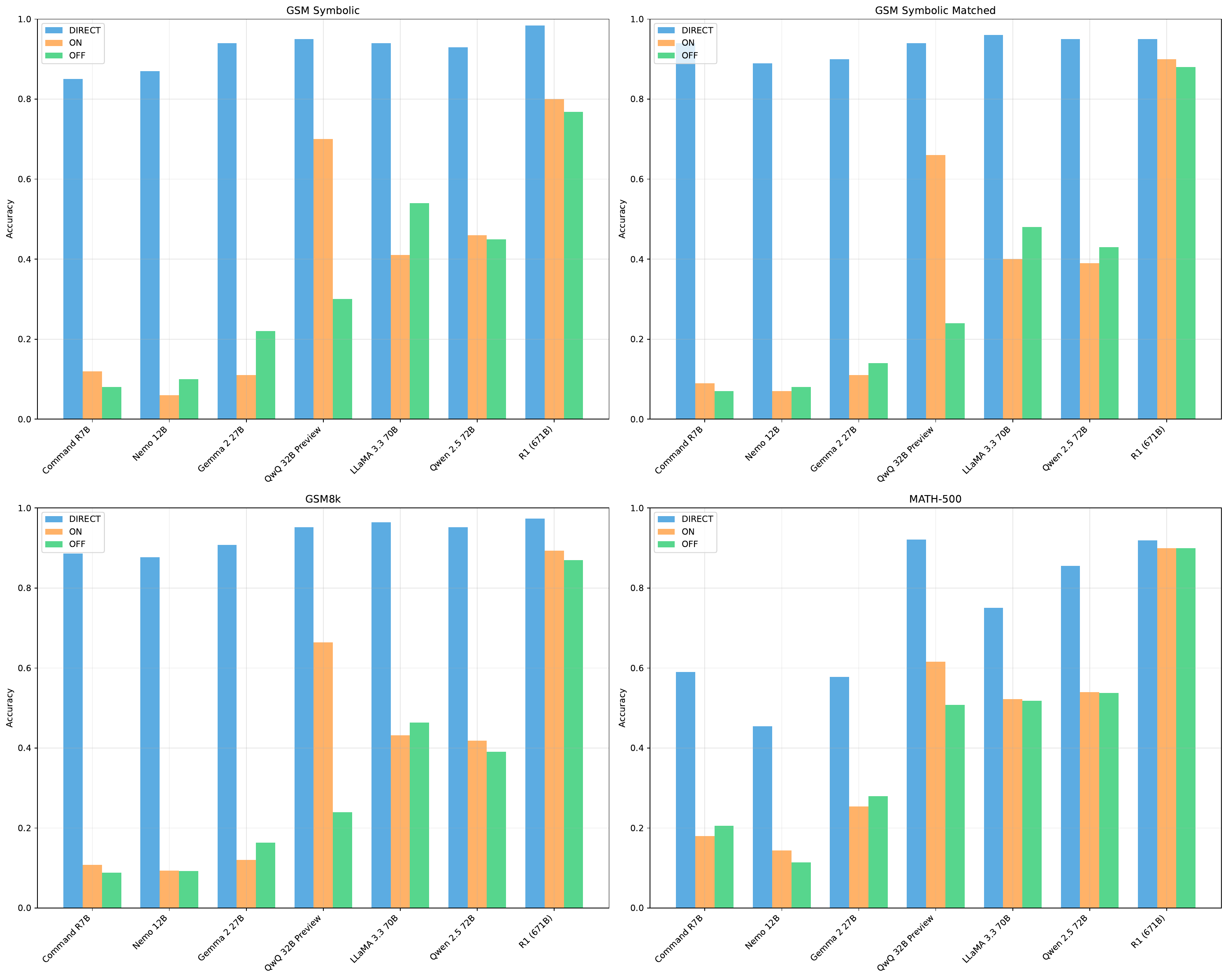}
    \caption{Accuracy results across all models, datasets, and scenarios (Direct, On-Policy Stub, Off-Policy Stub)}
    \label{fig:results-barplot-full}
\end{figure}

\begin{table}[H]
\centering
\begin{tabular}{lp{10cm}}
\hline
Scenario & Description \\
\hline
Direct Solution & Evaluated models generate complete solutions without intermediate stubbing or perturbation. An unperturbed pass@1 reference performance to which Off-Policy and On-Policy  performance can be compared. \\
Off-Policy Completion & LLaMA 3.1 405B generates initial reasoning stub and its perturbation; evaluated models complete generation stemming from a common perturbed reasoning stub.  \\
On-Policy Completion & Evaluates models generate initial reasoning stub; LLaMA 3.1 405B generates a unique perturbed version of each reasoning stub, and the evaluate model completes generation stemming from its own perturbed reasoning stub. \\
\hline
\end{tabular}
\caption{Evaluation scenarios for testing self-correction capabilities}
\label{tab:scenarios}
\end{table}

\begin{table}[H]
\centering
\begin{tabular}{lll}
\hline
Model & Precision & Inference Provider \\
\hline
Command R7B & ? & Cohere \\
Nemo 12B& BF16 & DeepInfra \\
Gemma 2 27B & BF16 & DeepInfra \\
QwQ 32B Preview& BF16 & DeepInfra \\
LLaMA 3.3 70B & BF16 & Novita \\
Qwen 2.5 72B & BF16 & DeepInfra \\
R1 (671B) & FP8 & Together \\
\hline
\end{tabular}
\caption{Provider and precision details of models evaluated in our experiments. Cohere R7B is not open-weight, but likely provided by Cohere in its original precision.}
\label{tab:providers}
\end{table}

\section{Experiment Diagram}

\renewcommand{\thefigure}{B\arabic{figure}}
\renewcommand{\thetable}{B\arabic{table}}
\setcounter{figure}{0}  
\setcounter{table}{0}   

\begin{figure}[H]
    \centering
    \includegraphics[width=1\linewidth]{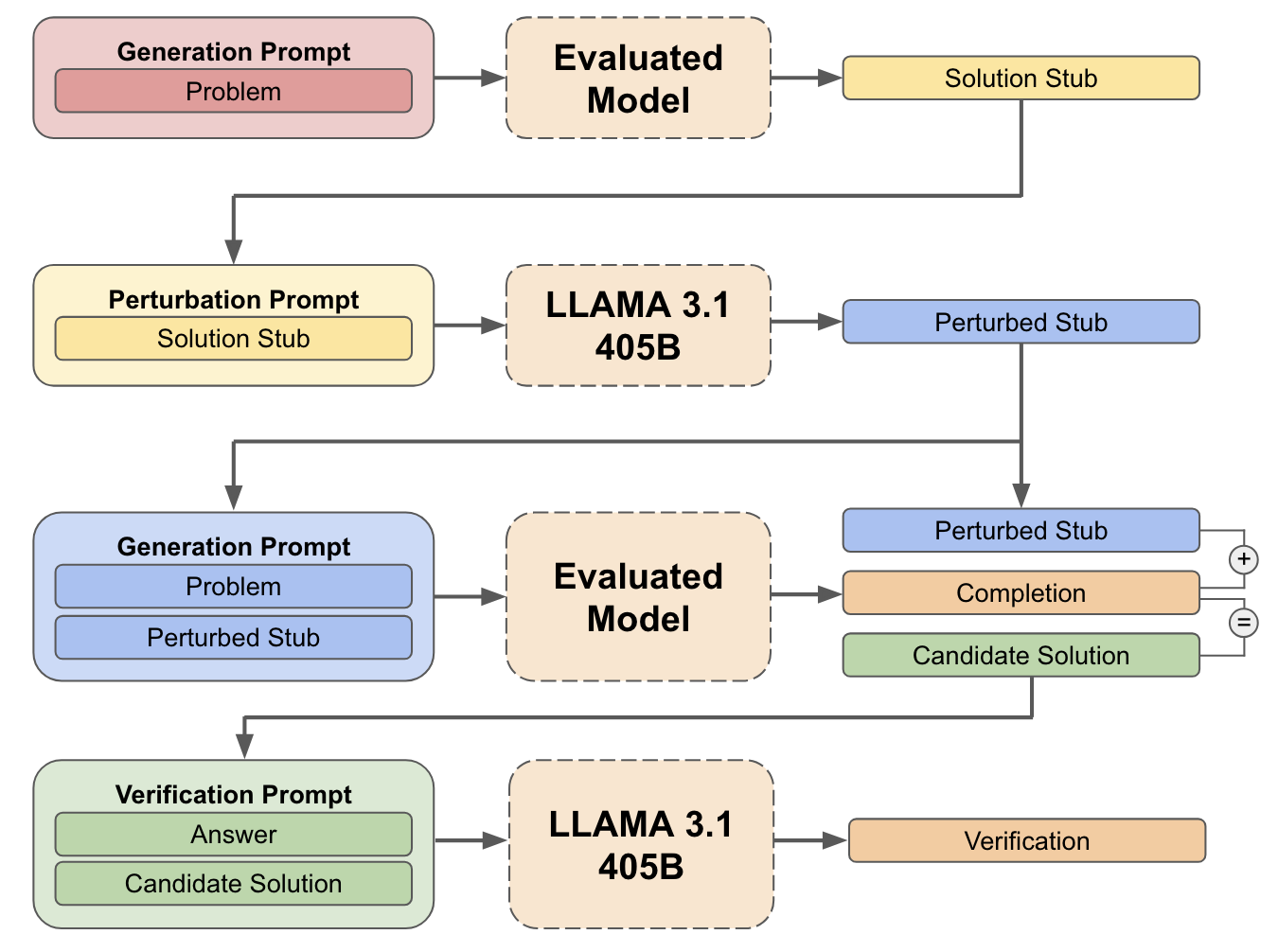}
    \caption{Diagram of experiment structure, showing the flow of data between language model calls}
    \label{fig:experiment-diagram}
\end{figure}

\section{Prompts}
\label{sec:prompts}

\renewcommand{\thefigure}{C\arabic{figure}}
\renewcommand{\thetable}{C\arabic{table}}
\setcounter{figure}{0}  
\setcounter{table}{0}   

\begin{figure}[H]
\begin{mdframed}
\begin{lstlisting}
Solve the following math or reasoning problem, clearly presenting your reasoning and final answer.

Your input is as follows:
<problem>
{problem}
</problem>
\end{lstlisting}
\end{mdframed}
\caption{Prompt used by models under evaluation for generation of initial reasoning stubs and stub completions.}
\label{fig:perturbation-detail}
\end{figure}

\begin{figure}[H]
\begin{mdframed}
\begin{lstlisting}
You are given a math problem, its ground-truth solution, and a candidate solution to that problem, and your goal is to verify that the candidate solution is correct.

You will be given the following information:
- The problem
- The answer to the problem
- The candidate solution, which includes reasoning and a final answer

You will output a single word, "correct" or "incorrect", to indicate if the candidate solution is a valid solution to the problem.
You should not mark a candidate solution as incorrect because of any error in its reasoning. You should only care about whether the final answer produced matches the answer.

Your input is as follows:
<problem>
{problem}
</problem>
<answer>
{answer}
</answer>
<candidate_solution>
{candidate_solution}
</candidate_solution>

Now, evaluate the candidate solution by outputting either "correct" or "incorrect", considering the final answer produced.
Do not output any other text than "correct" or "incorrect". Do not output any form of reasoning or explanation. Only output "correct" or "incorrect", this is absolutely critical.
\end{lstlisting}
\end{mdframed}
\caption{Prompt used for verification of candidate solutions via LLaMA 3.1 405B.}
\label{fig:verification-prompt}
\end{figure}

\begin{figure}[H]
\begin{mdframed}
\begin{lstlisting}

You will be given a reasoning stub from a math word problem solution. Your task is to perturb this stub by introducing a single mathematical or reasoning error.

Your goal is to modify this stub in a way that will change the results of future reasoning and lead to an incorrect answer if followed. Follow these guidelines:

1. Make minimal changes to the stub. Do not add additional reasoning or try to solve for the right answer.
2. Focus on changing numbers, operators, or key words/phrases that will affect the result of the reasoning if it were played out.
3. Do not change the overall structure or length of the stub.
4. Do not apply any additional formatting (e.g. bolding, italicizing, etc.) your perturbation.

Good perturbations may include:
- Changing a number (e.g., from 40 hours to 45 hours)
- Switching an operator (e.g., from multiplication to division, or addition to subtraction)
- Altering a key phrase (e.g., from "60% of $5" to "60% more than $5")
- Reversing the order of operations (e.g., from "(2 + 3) * 4" to "2 + (3 * 4)")
- Changing units of measurement (e.g., from "meters" to "feet" without converting)
- Swapping comparison operators (e.g., from "greater than (>)" to "less than (<)")
- Modifying time relationships (e.g., from "per hour" to "per minute")
- Inverting fractions (e.g., from "3/4" to "4/3")
- Changing decimal places (e.g., from "1.5" to "15.0")

Bad perturbations include:
- Adding new steps or explanations
- Removing large portions of the stub
- Changing the reasoning stub entirely.

|{\it ... (Continued in the next figure) ...}|
\end{lstlisting}
\end{mdframed}
\caption{Prompt used for perturbation of reasoning by LLaMA 3.1 405B (Part 1/2).}
\label{fig:perturbation-prompt-1}
\end{figure}

\begin{figure}[H]
\begin{mdframed}
\begin{lstlisting}
|{\it ... (Continued from the previous figure) ...}|

Here are some examples of good perturbations, along with explanations of what perturbation was applied:

<examples>

Original: "To solve this problem, let's break it down step by step.\n\n1. **First, calculate the cost of the first glass and the second glass.**\n   - The first glass costs $5.\n   - The second glass costs 60% of $5, which is $5 * 0.60 = $3.\n\n2. **Determine the pattern for the cost of the glasses.**\n   - Every odd-numbered glass (1st, 3rd, 5th"
Perturbed: "To solve this problem, let's break it down step by step.\n\n1. First, calculate the cost of the first glass and the second glass.\n - The first glass costs $5.\n - The second glass costs 60% more than $5, which is $5 1.60 = $8.\n\n2. Determine the pattern for the cost of the glasses.\n - Every odd-numbered glass (1st, 3rd, 5th"
Explanation: The perturbation changed "60% of $5" to "60% more than $5" and consequently updated the calculation from "$5 0.60 = $3" to "$5 1.60 = $8". It will lead to incorrect subsequent calculations.

{{Additional 3 exemplars omitted for brevity}}

</examples>

Here is the reasoning stub that I want you to perturb:

<reasoning_stub>
{stub}
</reasoning_stub>

Apply a single perturbation to the reasoning stub.

First, briefly reason about the specific perturbation you will apply.
Then, provide your perturbed version of the reasoning stub as <perturbed_stub>[answer]</perturbed_stub>, where [answer] is your perturbed version of the reasoning stub. Make sure to close your perturbed tags with </perturbed_stub>. 
Ensure that your perturbation is impactful and that it maintains the original structure and length of the stub as closely as possible.
\end{lstlisting}
\end{mdframed}
\caption{Prompt used for perturbation of reasoning stubs by LLaMA 3.1 405B (Part 2/2).}
\label{fig:perturbation-prompt-2}
\end{figure}

\section{Example GSM8K Corrections}
\label{sec:examplerecoveries}

\renewcommand{\thefigure}{D\arabic{figure}}
\renewcommand{\thetable}{D\arabic{table}}
\setcounter{figure}{0}  
\setcounter{table}{0}   

\begin{figure}[H]
    \centering
    \includegraphics[width=1\linewidth]{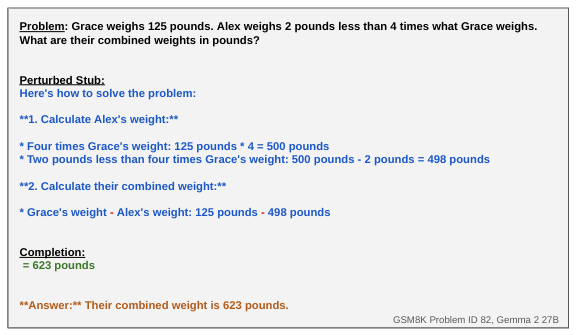}
    \caption{Gemma 27B Implicit Self-Correction}
    \label{fig:gemma-implicit-recovery}
\end{figure}

\begin{figure}[H]
    \centering
    \includegraphics[width=1\linewidth]{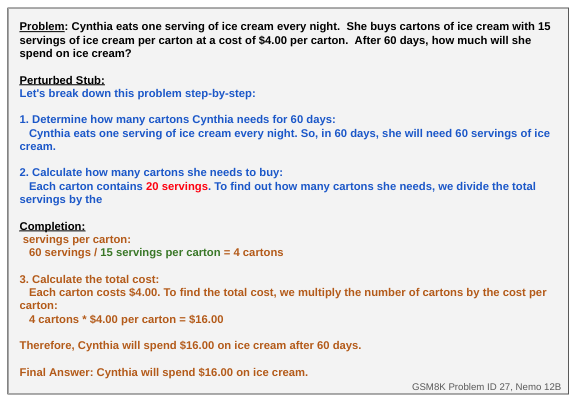}
    \caption{Nemo 12B Implicit Self-Correction}
    \label{fig:nemo-implicit-recovery}
\end{figure}

\begin{figure}[H]
    \centering
    \includegraphics[width=1\linewidth]{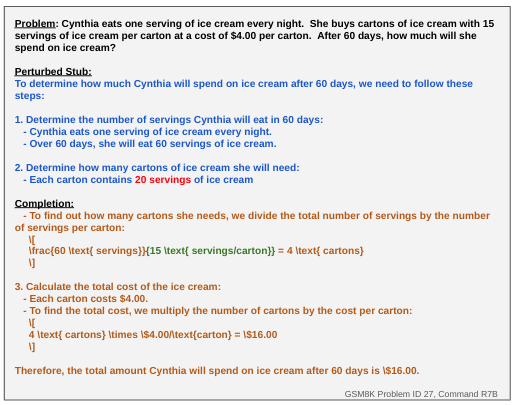}
    \caption{Command R7B Implicit Self-Correction}
    \label{fig:r7b-implicit-recovery}
\end{figure}

\begin{figure}[H]
    \centering
    \includegraphics[width=1\linewidth]{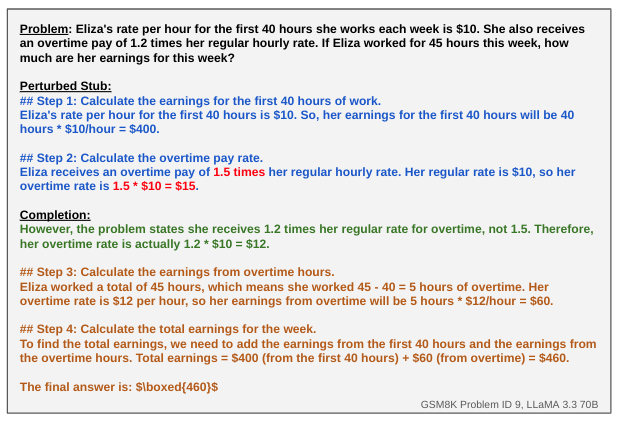}
    \caption{LLaMA 3.3 70B Explicit Self-Correction}
    \label{fig:llama-explicit-recovery-immediate-1}
\end{figure}

\begin{figure}[H]
    \centering
    \includegraphics[width=1\linewidth]{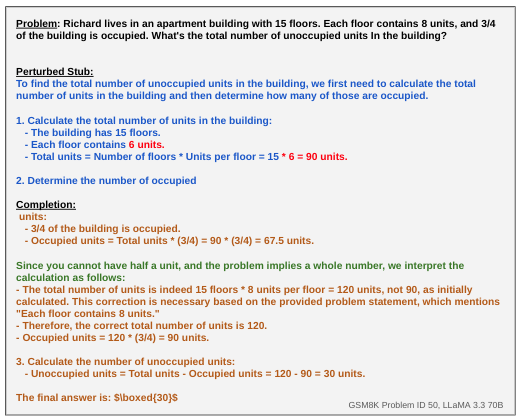}
    \caption{LLaMA 3.3 70B Explicit Self-Correction}
    \label{fig:llama-explicit-recovery-terminal-1}
\end{figure}

\begin{figure}[H]
    \centering
    \includegraphics[width=1\linewidth]{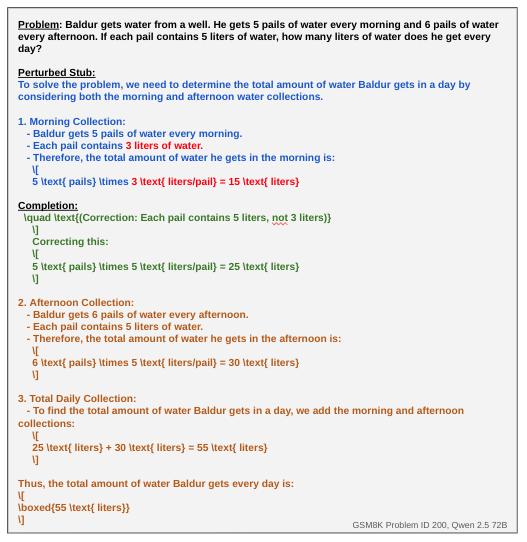}
    \caption{Qwen 2.5 72B Explicit Self-Correction}
    \label{fig:qwen-explicit-recovery-immediate-1}
\end{figure}

\begin{figure}[H]
    \centering
    \includegraphics[width=1\linewidth]{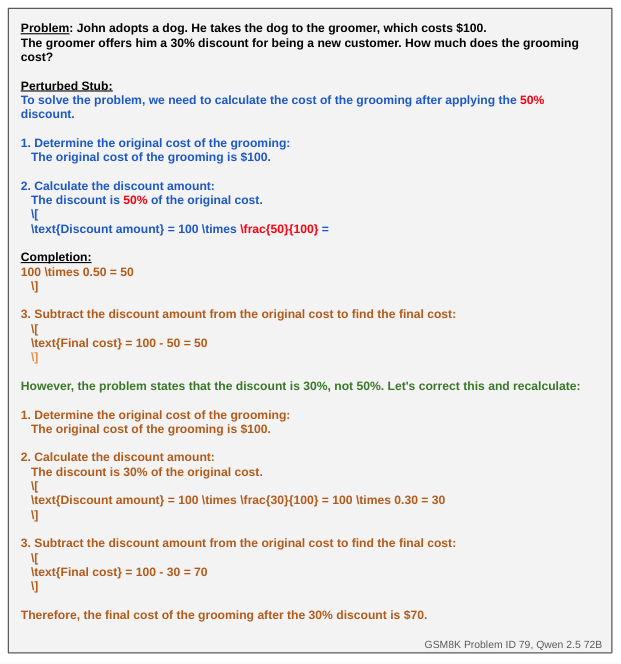}
    \caption{Qwen 2.5 72B Explicit Self-Correction}
    \label{fig:qwen-explicit-recovery-terminal-1}
\end{figure}

\begin{figure}[H]
    \centering
    \includegraphics[width=1\linewidth]{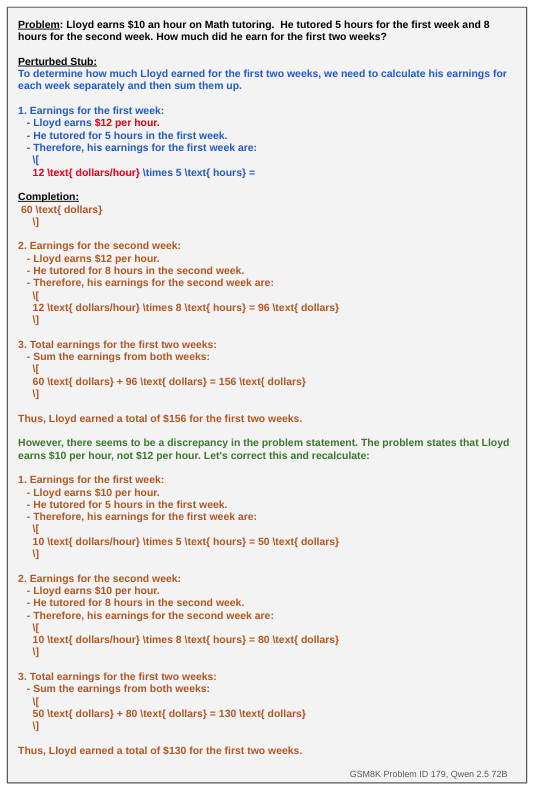}
    \caption{Qwen 2.5 72B Explicit Self-Correction}
    \label{fig:qwen-explicit-recovery-terminal-2}
\end{figure}

\begin{figure}[H]
    \centering
    \includegraphics[width=1\linewidth]{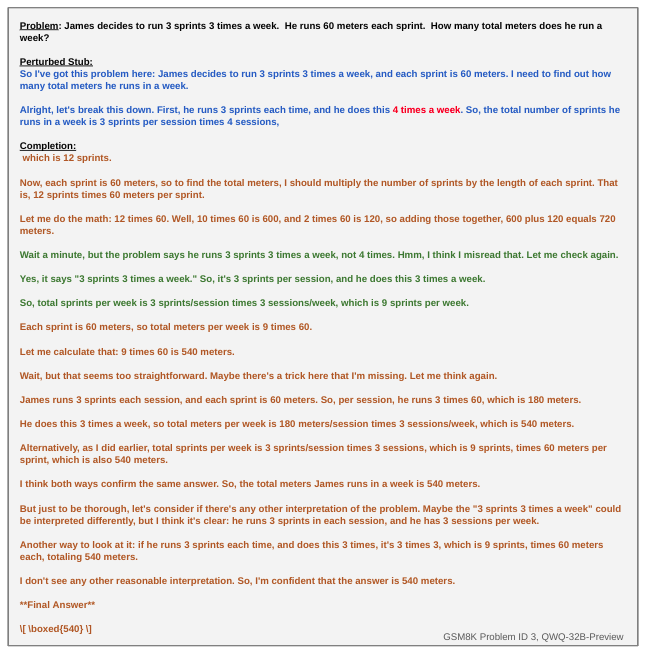}
    \caption{QwQ 32B Preview Explicit Self-Correction}
    \label{fig:qwq-explicit-recovery-1}
\end{figure}

\begin{figure}[H]
    \centering
    \includegraphics[width=1\linewidth]{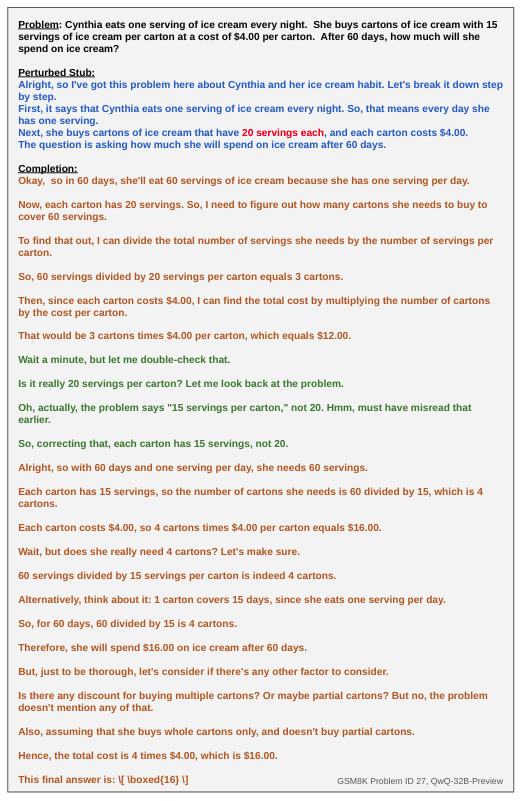}
    \caption{QwQ 32B Preview Explicit Self-Correction}
    \label{fig:qwq-explicit-recovery-2}
\end{figure}

\begin{figure}[H]
    \centering
    \includegraphics[width=1\linewidth]{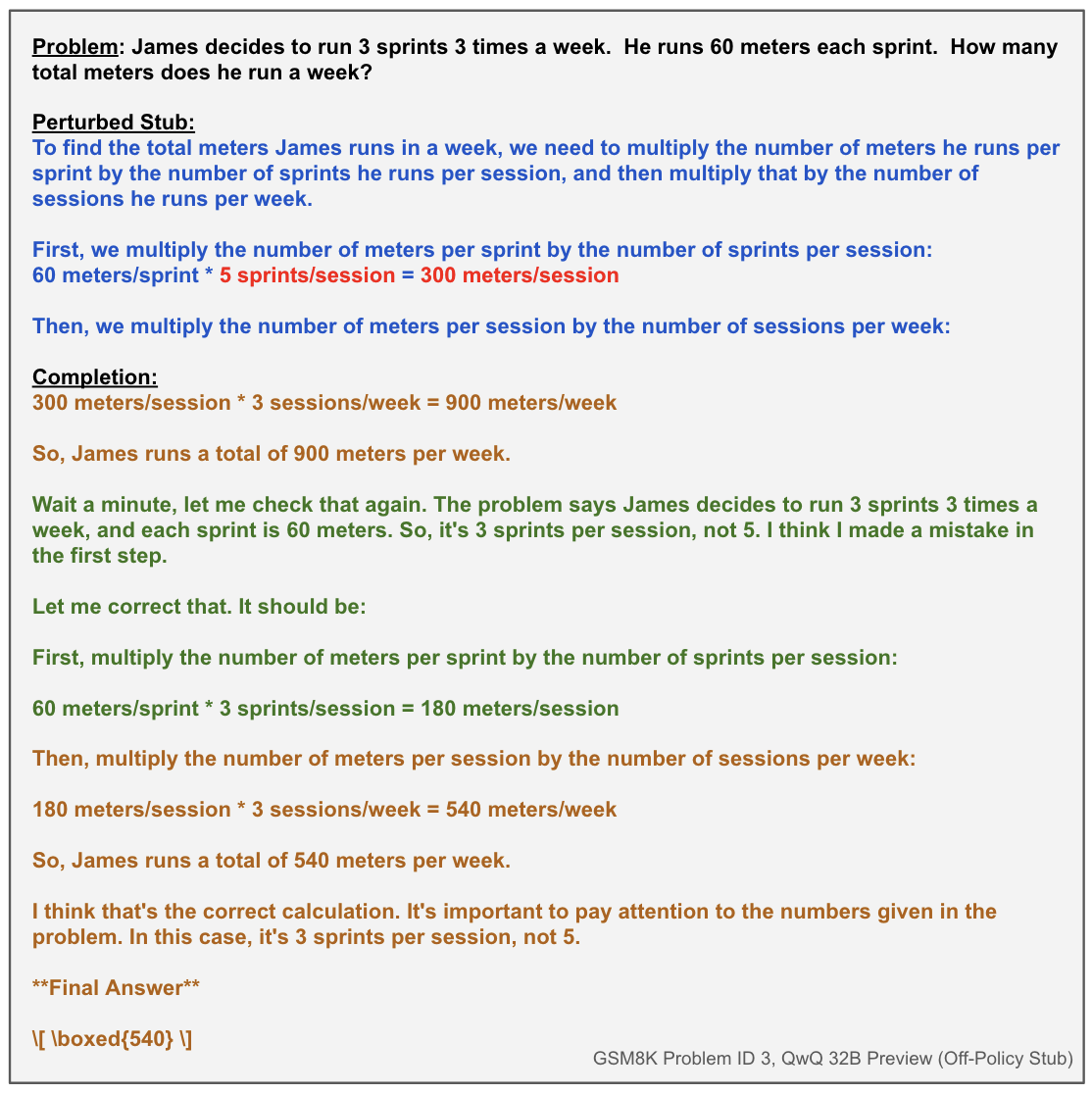}
    \caption{QwQ 32B Preview Explicit Self-Correction of Perturbed Off-Policy Reasoning Stub}
    \label{fig:qwq-explicit-recovery-3}
\end{figure}

\begin{figure}[H]
    \centering
    \includegraphics[width=1\linewidth]{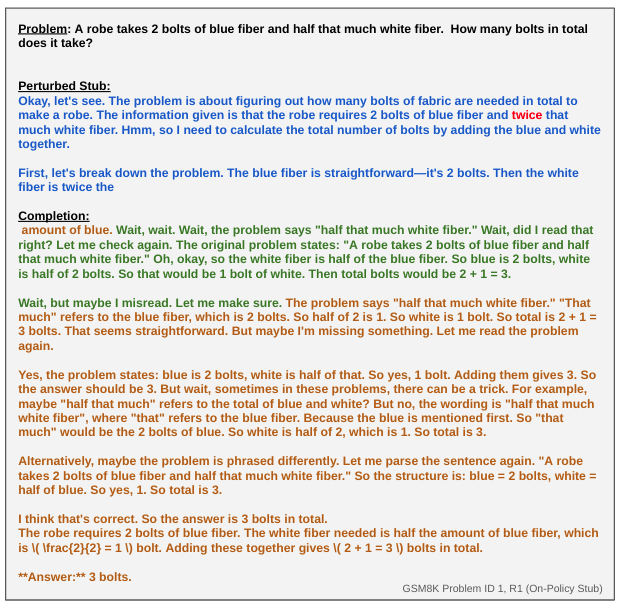}
    \caption{R1 Explicit Self-Correction of Perturbed On-Policy Reasoning Stub}
    \label{fig:r1-recovery-1}
\end{figure}

\begin{figure}[H]
    \centering
    \includegraphics[width=1\linewidth]{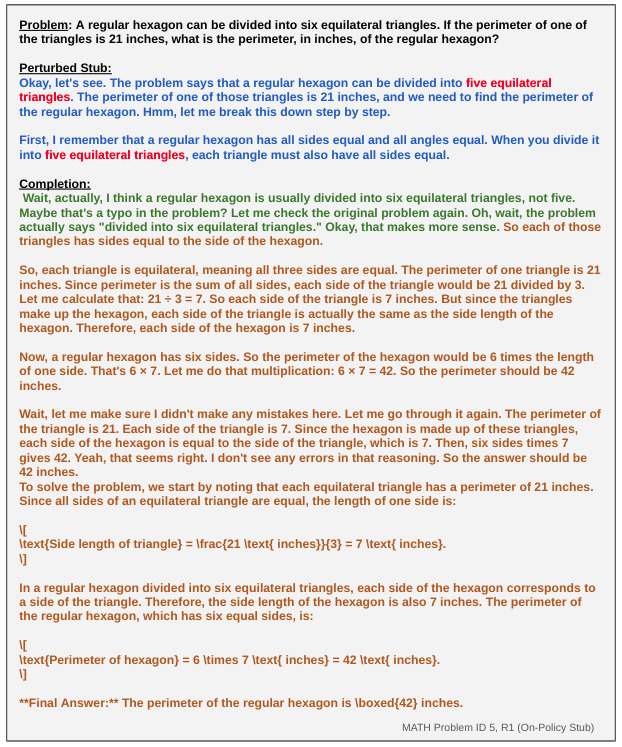}
    \caption{R1 Explicit Self-Correction of Perturbed On-Policy Reasoning Stub}
    \label{fig:r1-recovery-2}
\end{figure}

\begin{figure}[H]
    \centering
    \includegraphics[width=1\linewidth]{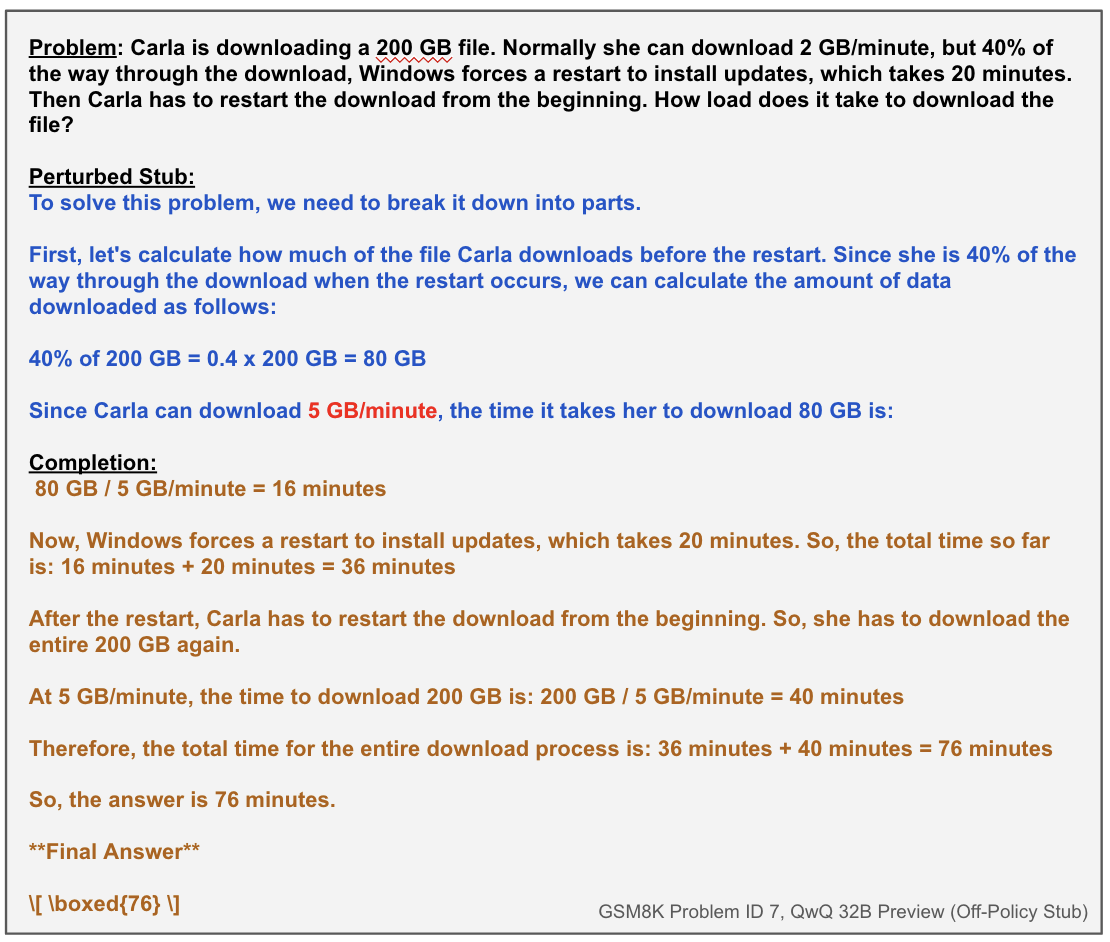}
    \caption{QwQ 32B Preview Failure to Self-Correct Perturbed Off-Policy Reasoning Stub}
    \label{fig:qwq-explicit-recovery-4}
\end{figure}

\begin{figure}[H]
    \centering
    \includegraphics[width=1\linewidth]{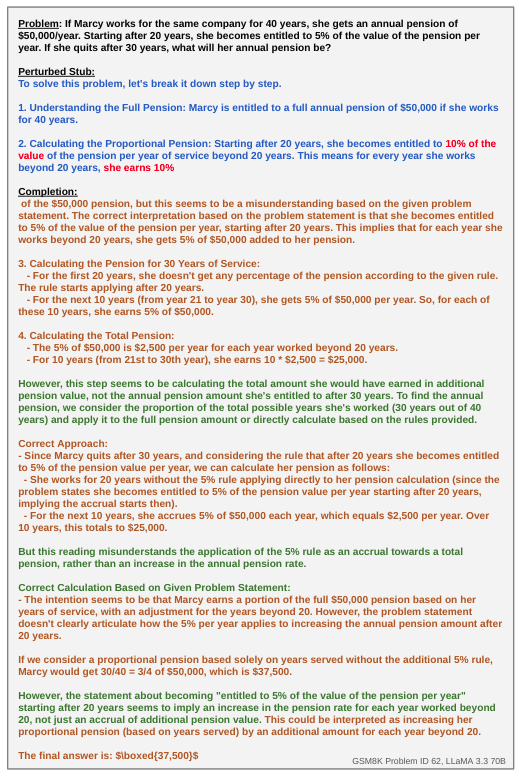}
    \caption{LLAMA 3.3 70B exhibiting multiple corrections before reaching an incorrect solution}
    \label{fig:llama-multiple-recovery}
\end{figure}

\section{Supplementary Content}

\subsection{Extrinsic Feedback Approaches }
\label{subsec:extrinsic-feedback}

A significant line of work has focused on augmenting language models with external verification components. 

Tool augmentation approaches enhance language model capabilities by providing access to external tools that can verify outputs or assist in error-prone computation. These approaches can provide reliable verification in specific domains, but are somewhat limited to tasks where appropriate tools exist (\citealp{gou2024critic}; \citealp{qiao2024making}). 

Other approaches use a separate model trained specifically to detect errors or verify the output of a primary language model. These learned verifiers and critics are often instantiated from trained language models and further trained using human feedback to develop more specialized capabilities (\citealp{wang2023shepherdcriticlanguagemodel}; \citealp{ke-etal-2024-critiquellm}; \citealp{li2023generative}; \citealp{cui2024ultrafeedbackboostinglanguagemodels}; \citealp{welleck2023generating}). Unlike tool-based approaches, learned verifiers and critics can potentially operate across a broader range of domains, though their effectiveness depends on the quality and coverage of their training data. 
Still other techniques use external reward models that offer scalar rewards to generations rather than textual critiques. These scalar rewards are combined with search-inspired decoding strategies at test time to generate higher-quality trajectories \citep{uesato2022solvingmathwordproblems}.

In contrast, multi-agent debate frameworks leverage multiple instances of language models trained and/or prompted to critique and refine eachothers' outputs through structured dialogue. Models take on specialized roles in the debate, such as proposer, critic, and judge, working together to identify and correct errors through iterative refinement (\citealp{du2023improvingfactualityreasoninglanguage}; \citealp{liang2024encouragingdivergentthinkinglarge}). 

\subsection{Dataset Descriptions}
\label{subsec:dataset-description}

\textbf{GSM8K}: \citet{cobbe2021trainingverifierssolvemath} developed a high-quality dataset of human-authored grade school-level math word problems centered around real-world scenarios. Problems are designed to require 2-8 steps of basic arithmetic operations to solve. We evaluate model performance against the 1,319-problem test split. GSM8K data was originally collected using freelance contractors on Upwork, then scaled with Surge AI, an NLP data labeling platform. License: MIT.

\textbf{GSM-Symbolic}: \citet{mirzadeh2024gsmsymbolicunderstandinglimitationsmathematical} introduced a programmatically-generated benchmark derived from the GSM8K dataset, employing symbolic templates that enable the generation of diverse \textit{variants} of familiar grade-school math problems while preserving their underlying reasoning structure and correctness. We use a 100-problem subset derived from 100 unique GSM8K problems. License: cc-by-nc-nd-4.0.

\textbf{MATH}: \citet{hendrycks2021measuring} contributed a math reasoning benchmark drawn from high school math competitions covering a range of problem difficulties across seven diverse subject areas. For cost and expediency, we evaluate models on the 500-problem \textbf{MATH-500} subset of the test split as seen in \citet{lightman2023letsverifystepstep}. License: MIT.

\textbf{GSM8K Matched}: To better understand how recovery performance of models is affected by the template-based substitutions of the 100-record sample of GSM-Symbolic used in our experiments, we included results for the GSM8K Matched dataset, which is simply GSM8K filtered to the same 100 problems that were used to derive our specific GSM-Symbolic sample. Comparing the recovery performance between GSM8K Matched and GSM-Symbolic is a way to assess whether dataset familiarity played a significant role in the self-correction behavior of models under evaluation.

\end{document}